\documentclass[times,twocolumn,final]{elsarticle}
\usepackage{medima}
\usepackage{framed,multirow}
\usepackage{amssymb}
\usepackage{latexsym}
\usepackage{medima}
\usepackage{url}
\usepackage{xcolor}

\usepackage{hyperref}

\usepackage{graphicx}
\usepackage{cite}
\usepackage{amsmath,amssymb,amsfonts}
\usepackage{algorithmic}
\usepackage{textcomp}
\usepackage{booktabs}
\usepackage{xspace}
\usepackage{svg}
\usepackage{amsmath}
\usepackage{footnote}
\usepackage{makecell}
\usepackage{multirow}
\usepackage{enumitem}
\usepackage{array}
\usepackage[nolist,nohyperlinks]{acronym}
\usepackage{subcaption}
\makesavenoteenv{tabular}
\usepackage{caption, subcaption}

\usepackage[acronym]{glossaries}
\makeglossaries
\acrodef{DL}{Deep learning}
\acrodef{CNN}{Convolutional Neural Network}
\acrodef{ReLU}{Rectified linear Unit}
\acrodef{mIoU}{mean Intersection over Union}
\acrodef{ROI}{Region of Interest}
\acrodef{AI}{Artificial Intelligence} 
\acrodef{CADx}{computer aided diagnosis} 
\acrodef{CRC}{Colorectal Cancer}
\acrodef{SGD}{Stochastic Gradient Descent}
\acrodef{DSC}{Dice Coefficient}
\acrodef{IoU}{Intersection over Union}
\acrodef{MAML}{Model Agnostic Meta Learning}
\acrodef{iMAML}{Implicit Model Agnostic Meta Learning}
\acrodef{MRI}{Magnetic Resonance Imaging}
\acrodef{ML}{Machine Learning}
\acrodef{PMG}{Prior-mask-guided}

\definecolor{newcolor}{rgb}{.8,.349,.1}

\journal{Computers in Biology and Medicine
}

\begin{document}

\verso{R.Khadka \textit{et~al.}}

\begin{frontmatter}


\title{Meta-learning with implicit gradients in a few-shot setting for medical image segmentation}%
\author[1,4]{Rabindra Khadka}
\author[1,2]{Debesh Jha}
\cortext[cor1]{Corresponding authors}\ead{debesh@simula.no}
\author[1,4]{Steven Hicks}
\author[1,4]{Vajira Thambawita}
\author[1,2]{Michael A. Riegler}
\author[3,5,6]{Sharib Ali}\ead{sharib.ali@eng.ox.ac.uk}
\author[1,4,6]{P{\aa}l Halvorsen}  
\address[1]{SimulaMet, Oslo, Norway}
\address[2]{UiT The Arctic University of Norway, Troms{\o}, Norway}
\address[3]{Department of Engineering Science, Institute of Biomedical Engineering, University of Oxford, Oxford, UK}
\address[4]{Oslo Metropolitan University, Oslo, Norway}
\address[5]{NIHR Oxford Biomedical Research Centre, University of Oxford, Oxford, UK}
\address[6]{shared senior authorship}
\received{x xxxx xxxx}
\finalform{x xxxx xxxx}
\accepted{x xxxx xxxx}
\availableonline{x xxxx xxxx}
\communicated{xxxx xxxx}

\begin{abstract}
Widely used {traditional} supervised deep learning methods require a large number of training samples but often fail to generalize on unseen datasets. {Therefore}, a more general application of any trained model is {quite} limited for medical imaging for clinical practice. Using separately trained models for each unique lesion category or a unique patient population will require {sufficiently} large curated datasets{,} which is not practical to use in {a real-world} clinical {set-up}. Few-shot learning approaches can not only minimize the need for {an} enormous number of reliable ground truth labels that are {labour-intensive and expensive}, but can also be used to model on a dataset coming from a new population. To this end, we propose to exploit an optimization-based implicit model agnostic meta-learning (iMAML) algorithm under few-shot settings for medical image segmentation. Our approach can leverage the learned weights from diverse but small training samples to perform analysis on unseen datasets with high accuracy. We show that, unlike classical few-shot learning approaches, our method {improves} generalization capability. To our knowledge, this is the first work that exploits iMAML for medical image segmentation and explores the strength of the model on scenarios such as meta-training on unique and mixed instances of lesion datasets. Our quantitative results on publicly available skin and polyp datasets show that the proposed method outperforms the naive supervised baseline model and two recent few-shot segmentation approaches by large margins. In addition, our iMAML approach shows an improvement of 2\%-4\% in dice score compared to its counterpart MAML for most experiments.
\end{abstract}

\end{frontmatter}
\section{Introduction}\label{sec:introduction}
Automatic lesion segmentation can help in accurate quantification of the area covered by anomalies, precise surgical removal, and treatment. Unlike manual processes, which are usually subjective and sub-optimal, automated methods can provide a more objective analysis of the lesions and their risks. \ac{ML} and \ac{DL}-based models have already shown successful results in the clinical settings~\citep{horng2017creating,brown2018automated,hinton2018deep}.

{Data shift and availability of labeled data are major bottlenecks in medical image analysis. Other challenges that medical image analysis has to deal with are: 1) difficulty to get domain experts to perform annotations, 2) heterogeneous data, e.g., it could consist of multiple organs (skin, gastrointestinal organs) and varied disease types (melanoma in skin and polyp in the colon), and 3) large variability between expert and novice annotations. The lack of publicly available datasets as well as their quality (e.g., missing and erroneous labels)  pose additional challenges. In addition, widely used supervised deep learning approaches require large amounts of training samples with labels and often fail to generalize when tested on different datasets due to data shifts caused by different data distribution. Data shift can arise due to population variation (e.g., different demographics), acquisition difference (e.g., devices or imaging protocols), prevalence shift (e.g., environmental factors affecting organs) and selection bias (e.g., inclusion criteria for study)~\citep{castro2020causality}.} 

The state-of-the-art \ac{DL} models require a large number of high-quality and diverse datasets with pixel-wise masks for {segmentation} that {is} difficult to generate. Additionally, publicly available datasets are still limited and often include only a few samples of each unique class, case or part of a population. Some example datasets include KvasirCapsule-SEG~\citep{jha2021nanonet} (55 samples), ETIS-Larib~\citep{silva2014toward} (196 samples), PH\textsuperscript{2}\citep{mendoncya2013dermoscopic} (200 samples), EDD2020~\citep{ali2020endoscopy} (386 samples),
Kvasir-instrument~\citep{jha2021kvasir} (590 samples), and CVCClinicDB~\citep{bernal2015wm} (612 samples). With the available datasets, it is still possible to build a \ac{ML} model by leveraging semi-supervised or few-shot learning methods~\citep{feyjie2020semi}, but these datasets listed above do not cover all lesion categories or include data from multiple sources; for example, rare disease cases, patient variabilities and multi-center data sources. Therefore, it is challenging to design a model that generalizes well on unseen datasets during clinical deployment. The possible solutions to the dataset mismatch can be domain adaptation~\citep{ganin2016domain} and domain generalization~\citep{li2017deeper}. Domain adaptation utilizes a labeled source training dataset and unlabeled target data to develop a model that performs well on the target environment. Implementing such adaptation techniques helps to increase the generalization capacity of the model towards unseen target domain configuration. On the other hand, domain generalization capitalizes on using multi-source training datasets to design a classifier that generalizes well on unseen target (test) datasets. However, the problem of data scarcity is not resolved by any of these techniques in a classical supervised setting as they require large training datasets~\citep{dou2019domain,Ghifary_2015_ICCV,motiian2017unified}. In addition, in domain adaptation methods, the learnt features have a similar embedding for both source and target dataset, and hence, this trade-off leads to compromises in the generalization capacity of the model.~\citep{tsai2018learning,celik2021endouda}.

To mitigate the problem of data scarcity and domain generalization, meta-learning under few-shot settings has emerged as a potential solution~\citep{ravi2016optimization,finn2017model}, especially in limited data settings. Meta-learning enables learning model weights by leveraging prior knowledge from various tasks~\citep{thrun2012learning} and can be implemented in different task objectives such as few-shot learning or multi-task learning. It is advantageous to use meta-learning in few-shot settings, and it has been primarily used in image classification~\citep{ali2020additive,mahajan2020meta}. Few-shot learning is a method that uses few annotated examples (support set) to make predictions on unlabeled examples (query set) and is the most appropriate choice when only limited data samples are available. An episodic training in a meta-learning setting can exploit to generalize to such limited data settings and become a natural choice for other tasks such as segmentation. Few-shot learning for segmentation has mostly been explored for natural images~\citep{zhang2020sg,zhang2019canet}. Recently, it is gaining more attention in the medical image segmentation~\citep{khandelwal2020domain,feyjie2020semi,rutter2019convolutional,liu2020shape,zhang2021domain,khandelwal2020domain,xiao2021prior}. Recent work by~\citet{feyjie2020semi} used a semi-supervised few-shot learning approach to perform skin lesion segmentation by feeding the learner with unlabeled surrogate tasks. \citet{roy2020squeeze} applied a few-shot technique with a squeeze and excite block architecture to perform volumetric segmentation of multiple organs in medical images. In  {the} work proposed by~\citet{ouyang2020self}, few-shot segmentation with a self-supervised method has been used to eliminate the need for having annotated medical images. They used an adaptive local pooling module in conjunction with prototypical networks to perform segmentation.

{{There are also some studies done to address the data scarcity and data mismatch problems in medical imaging field based on Wasserstein generative adversarial networks (GAN) where it was adopted for image reconstruction~\citep{LowDoseCI,GAN_Circle,DentalInlay}. {Some studies have been carried out} to mitigate the generalization problem of the \ac{ML} model in the medical domain, like the work done by~\citep{cervical} where they developed multi-scale deep convolutional networks that {perform} segmentation of overlapping cytoplasm. The work done by~\citep{overlapping_bacteria} proposes {an} automatic method to segment overlapping bacteria regions where they also incorporate Markov random field for unsupervised segmentation of small {objects}. These methods show improved generalization capacity. {However,} these methods are not tested under {a} few-shot setting.}} Furthermore, {all} of these works based on the few-shot learning approach use the same data source and similar tasks for inference, which means {that} the data shift problem has not been tackled.

\begin{table*}[!t]
 \caption{Publicly available medical imaging datasets used in our experiments. {Here we provide the number of image samples, size of images and imaging type that were incorporated in these datasets.}}
    \label{table:datasettable}
    \centering
         \begin{tabular}{ l c c c c}  
        \toprule
        \textbf{Dataset} & \# of \textbf{Images} & \textbf{Input size}  & \textbf{Imaging type} \\ 
        \midrule
        Kvasir-SEG~\citep{jha2020kvasir}  & 1000 & Variable &  Colonoscopy & 
        \\  
        KvasirCapsule-SEG~\citep{jha2021nanonet}  & 55 & Variable &  Video capsule endoscopy & 
        \\  
        CVC-ClinicDB~\citep{bernal2015wm} & 612 & $384\times 288$ &  Colonoscopy \\  
        \shortstack{ISIC-2018~\citep{codella2019skin,tschandl2018ham10000}}& 2596 & $384 \times 512$ & Dermoscopy  \\ 
        PH\textsuperscript{2}~\citep{mendoncya2013dermoscopic}& 200 & $768 \times 560$ & Dermoscopy \\ 
        \bottomrule
\end{tabular}
\end{table*}

A recent study~\citep{oliver2018realistic} suggested that the supervised transfer learning method with fine-tuning can handle the data mismatch better than semi-supervised methods.
{The few-shot semi-supervised method adopted by~\citet{feyjie2020semi} does not show {a} promising result, the {predicted} mask stands just at 62.40\% of the target mask (ISIC dataset) under {the} 5-shot setting.} Thus, in our work{,} meta-learning is adopted for domain generalization by further optimizing model weights via a meta-optimizer to overcome the shortcomings of few-shot learning. Recent work by~\citet{dou2019domain} used the gradient-based meta-learning algorithm known as \ac{MAML}, where the idea was to operate in the semantic feature space and learn semantically invariant features across training domains. They evaluated their method with \ac{MRI} images of the brain from different datasets that inherited domain shifts. They showed consistent results across all the datasets. However, the approach has not been tested under few-shot settings, i.e., less number of samples given during training to adapt to generalization capability in resource constraint settings during inference. Also, the training and test {set included} instances from the same anatomy. Being able to generalize well over another lesion type by training on one lesion type can be advantageous in medical imaging to tackle data scarcity problems. Additionally, the used \ac{MAML} algorithm by~\citet{dou2019domain} has some caveats related to computation and memory efficiency, which makes it difficult to scale up the accuracy as it requires several optimization steps~\citep{rajeswaran2019meta}. The \ac{iMAML} algorithm~\citep{rajeswaran2019meta,khadkameta}, on the other hand, can provide faster and improved optimization during the meta-learning 
since the solution depends only upon the inner optimization and not the path taken by {an} inner optimization algorithm.

Primarily, this work explores the efficacy of the iMAML algorithm for medical image segmentation with the
objective of handling the problem of data scarcity and data shift. We propose to demonstrate the use of iMAML in {medical} image segmentation and compare the results with other semi-supervised approaches. During this study, the convexity of the dice loss function is improved by applying Lov\`{a}sz extension~\citep{Lovsz}. We also compared the \ac{iMAML} algorithm with the \ac{MAML} algorithm under the same setting. 
The requirement of {a} few-shot learning paradigm to tackle data limitations is well established. However, the {fine-tuning} of the weight parameters has been revisited in several studies showing the ability of the model agnostic meta-learning approach. To this end, no studies have used {an} implicit model agnostic approach for medical image segmentation.
Our contribution includes (i) incorporation of attention-UNet~\citep{oktay2018attention} mechanism for inner optimization of the weights using segmentation tasks on two different datasets during episodic meta-training, (ii) utilizing an analytical solution (conjugate gradient) for computing meta-gradients to achieve optimized weights, and (iii) a comprehensive analysis of the efficacy of \ac{iMAML} on publicly available skin and polyp datasets from multiple sources.
{Our paper is structured into {the} following sections: Section 2 details the datasets used in this work{,} in Section 3, we present our iMAML segmentation approach and the compound loss function, Section 4 contains the experimental details and results, in Section 5 we provide comprehensive discussion and finally in Section 6 we conclude the paper}.

\section{Datasets}
We use five widely used publicly available datasets, namely
Kvasir-SEG~\citep{jha2020kvasir}, KvasirCapsule-SEG~\citep{jha2021nanonet},
CVC-612~\citep{bernal2015wm}, 
ISIC-2018~\citep{codella2019skin,tschandl2018ham10000}, and
PH\textsuperscript{2}~\citep{mendoncya2013dermoscopic}. A combination of these datasets has been used for the meta-training stage and tested on a holdout dataset to evaluate our proposed iMAML segmentation approach. Table~\ref{table:datasettable} presents information of each dataset used in our experimental setup. 

Kvasir-SEG~\citep{jha2020kvasir} is a widely used publicly available colonoscopy dataset. It consists of 1000 polyp images, their corresponding ground truth segmentation masks and bounding boxes information of the area covered by polyp. The example images from the Kvasir-SEG dataset can be found in Figure~\ref{fig:qualitative}.  The size of each polyp varies from $332\times 487$ to $1920\times 1072$. The original images from the Kvasir-SEG are captured during a colonoscopy examination using the ScopeGuide colonoscope (Olympus). The dataset can be downloaded from \url{https://datasets.simula.no/kvasir-seg/}.

\begin{figure}
    \centering
    \includegraphics[width = 9cm] {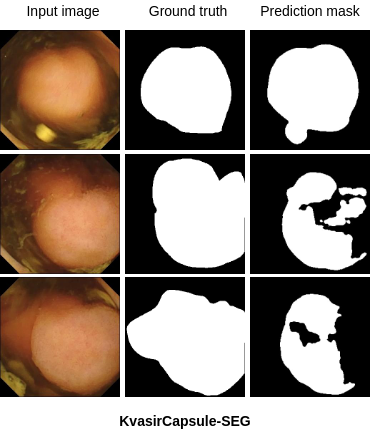}
    \caption{Qualitative results on the KvasirCapsule-SEG dataset {corresponding to Table~\ref{tab6}. The illustrated prediction maps (left) refer to the results from our third setup where samples comprised of two different datasets but with unique class (i.e., skin datasets, PH$^2$ and ISIC-2018, in this case).}} 
    \label{fig:kvasircapsule}
\end{figure}

KvasirCapsule-SEG~\citep{jha2021nanonet} is the wireless video capsule endoscopy dataset. This dataset was developed by annotating the ground truth segmentation maps from the polyp images found in the Kvasir-Capsule dataset~\citep{smedsrud2021kvasir}. The dataset consists of 55 polyp images and their corresponding ground truth segmentation masks and bounding boxes. The example of KvasirCapsule-SEG can be found in Figure~\ref{fig:kvasircapsule}. The dataset can be downloaded from \url{https://www.kaggle.com/debeshjha1/kvasircapsuleseg}.

CVC-ClinicDB~\citep{bernal2015wm}, also known as CVC-612, is another popular polyp segmentation dataset. It consists of 612 polyp images from 31 colonoscopy videos and their corresponding ground truth masks. The sample images from CVC-ClinicDB can be found in {stage} 1 of the Figure~\ref{fig:method}. The dataset is available at \url{https://www.dropbox.com/s/khtlmehjgv1b07z/cvc612.zip?dl=0}.

The ISIC-2018 dataset~\citep{codella2019skin,tschandl2018ham10000} includes both benign and malignant skin lesion images. It consists of 2596 dermoscopy images and their corresponding ground truth masks. The example samples can be observed in Figure~\ref{fig:qualitative}. The image resolution is $384\times 512$, and the dataset can be downloaded from \url{https://challenge.isic-archive.com/data}. 

The PH\textsuperscript{2}\citep{mendoncya2013dermoscopic} dataset consists of dermoscopic images. It consists of 200 images of melanocytic lesions. The ground truth segmentation mask for each image is provided. The dataset can be downloaded from \url{https://www.dropbox.com/s/k88qukc20ljnbuo/PH2Dataset.rar}. More details about the dataset can be found on the webpage\footnote{\url{https://www.fc.up.pt/addi/ph2\%20database.html}}. 

\section{Methodology}
{This section describes the algorithm and the adopted method used to obtain the empirical results.}  

\subsection{iMAML algorithm}
\begin{figure*}[t!]
    \centering
    \includegraphics[width=0.98\textwidth]{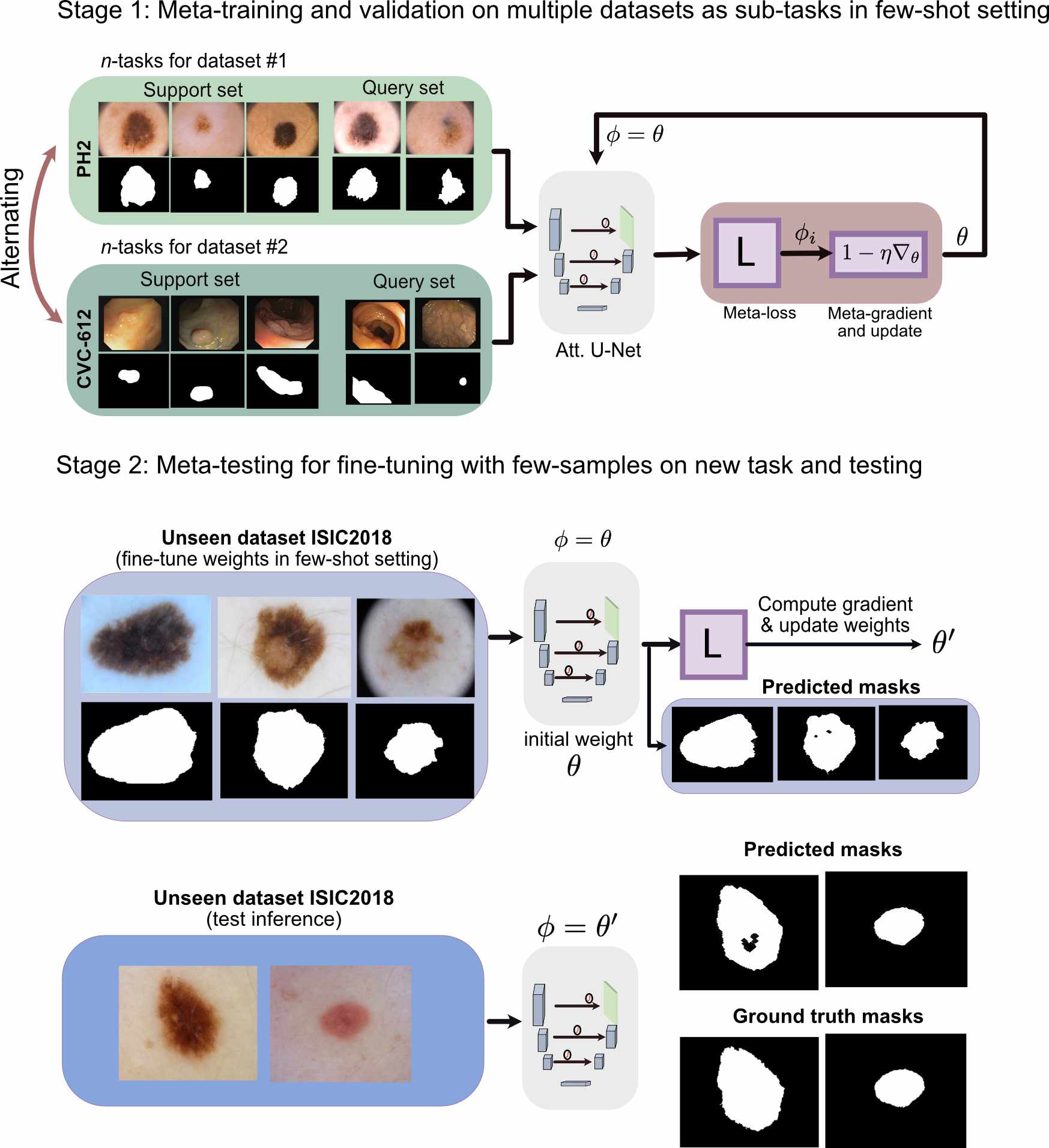}
    \caption{\textbf{Meta-learning with {an} implicit gradient optimization on medical imaging datasets}: Meta training is done as episodic tasks on two public datasets (\#1 and \#2). In the first stage, a few-shot learning framework for each task is used for the support set, and validation is done on the query set. During the meta-testing stage, an unseen task from the third dataset is provided with the optimized weights obtained from the first stage, \#1 and the gradient of the computed loss is used to readjust the final weights on only few samples of this dataset. Finally, the fine-tuned weight is used for the inference of the test samples. In all these {settings}, we use attention U-Net~\citep{oktay2018attention} to achieve segmentation maps.}
    \label{fig:method}
\end{figure*}
 
In general, \ac{MAML} approaches are trained through a meta-learning objective function~\citep{finn2017model}. However, due to the requirement of back-propagation during model training with high-order meta-gradients, \ac{MAML} can suffer from vanishing gradients. In order to eliminate this problem, \citet{rajeswaran2019meta} suggested to use a bi-level optimization, where an \textit{inner} optimization is focused on computing weights through the \ac{CNN} model and an analytic solution is used for the \textit{outer} meta-gradient estimation (see Eq.~(\ref{eq:1})).

\begin{equation}\label{eq:1}
\begin{split}
\theta^{*}_{ML}:\equiv argmin_{\theta} \underbrace{\frac{1}{M} \sum^{M}_{i=1} \ L(\overbrace{Alg_i(\theta,D^{tr}_{i})}^{inner-level},D^{val}_{i})}_{outer-level},\quad \text{with}\\
 Alg_{i}(\theta):= \underset{\phi \in \theta} {argmin} \ L_{i}(\phi)+\frac{\lambda}{2}||\phi-\theta||^{2}
\end{split}
\end{equation}

 In Eq.~\ref{eq:1}, $D_i^{tr}$ and $D_i^{val}$ represent training (support set) and validation (query set) in the meta-training phase for the $i^{th}$ task. The task-specific parameters in the inner optimization level are represented by $\phi$ while the optimized weights after meta-training, i.e., the meta-parameters, are represented by $\theta$. The final optimized meta-parameters are represented as $\theta^{*}_{ML}$. In order to avoid overfitting and help anchor, the task parameter $\phi$ to the meta-parameter $\theta$, an L2-regularization is used for the model training $Alg_i$. 

The meta-training and meta-testing stages are shown in Figure.~\ref{fig:method}. During the meta-training stage, tasks are generated. The tasks contain {a} support set (train) and query set (validation) with few-shot instances. This means that only a few samples are chosen, such as 5 for 5-shot and 10 for 10-shot. We then initialize our attention U-Net segmentation model with random weights $\theta_o$ for the $i^{th}$ task. We then computed the loss $L$ between the predicted mask and the ground truth mask in the support set with $L2$ -regularization. Validation loss on the query data completes the task for which the optimized $\phi_i$ is fed to the meta-learner where meta-gradients are analytically computed and updated as in Eq.~(\ref{eq:2}). This is then fed to the model weights of the attention U-Net architecture for further backpropagation and optimization. Such a two-level optimization scheme is iterative and done for two different datasets in our case (see Fig.~\ref{fig:method}, top). The meta-training stage is completed once the set number of tasks $M$ are completed to obtain the final meta-learned parameters $\theta^{*}_{ML}$. 

\begin{equation} \label{eq:2}
  \theta \  \leftarrow \ \theta- \eta\frac{1}{M} \sum^{M}_{i=1} \frac{d \ Alg_{i}(\theta)}{d\theta} \ \nabla_{\phi}\ L_{i}(Alg_{i}(\theta))
\end{equation}

The second stage consists of a simple fine-tuning step on the unseen data where optimized weight $\theta^{*}_{ML}$, say $\theta$ for simplicity, is used to optimize the loss function $L$ in a few-shot setting. The {resulting} final weights are then used in the final inference for direct segmentation map prediction as shown in Fig.~\ref{fig:method} (bottom). 

\subsection{Loss function}
A compound loss was used during training which comprises of both \emph{log-cosh-dice loss} and \emph{binary cross entropy loss}. { It attenuates the problem of class imbalance through dice-loss. } The final loss function is devised as:
\begin{equation}
\begin{split}
L = L_{BCE}+ L_{lc-dce} +\lambda||\theta||^{2}_{2}, \quad {\text{with}}\\
  L_{lc-dce}=log  ( cosh (L_{Dice}) )
 \end{split}
\end{equation}
where;
\begin{equation} \label{eq:4}
\begin{split}
  {L_{BCE}=-(y \log( \hat{y})+(1-y)\log(1-\hat{y}))}
\end{split}
\end{equation}

\begin{equation} \label{eq:5}
\begin{split}
{L_{Dice}=1-\frac{(2 \sum_{i}y_{i} \hat{y_{i}})+1} {\sum_{i}y_{i}+\sum_{i}\hat{y_{i}}+1}}
\end{split}
\end{equation}
Here, 
{{$\hat{y_{i}}$ and $y_{i}$ refer to  the pairs of corresponding pixel values of prediction and  ground-truth, respectively.}}

\noindent{$L_{Dice}$} and $L_{BCE}$ have usual meanings for dice loss and binary cross-entropy loss classically used in segmentation approaches~\citep{cosh}. {Binary Cross-entropy~\citep{BCE} quantifies the difference between two probability distributions for a given random variable (eqn~\ref{eq:4}). It is popularly adopted for object classification or pixel-level classification during segmentation. Dice loss ({$L_{Dice}$})~\citep{SudreLVOC17} is based on dice coefficient, which measures the overlap between predicted and ground-truth masks (eqn~\ref{eq:5})}.  Unlike classical dice loss, $L_{lc-dce}$ is the Lov\`{a}sz extension~\citep{Lovsz} that tackles the non-convex nature of dice loss by smoothing it and making the function tractable and easy to differentiate. Additionally, we have added a weight decay function as an $L_{2}$ regularization with $\lambda$ as regularization hyper-parameter, and $\theta$ is the model weight. This allows to encapsulate better generalizability on test samples.

\begin{table*}[!t]
    \centering
     \caption {{Quantitative results as DSC metric for our first experimental setup.} Here, episodic training for meta-learning is done independently with 50 tasks, first on PH\textsuperscript{2} (skin) and then on Kvasir-SEG (polyp). { Here, naive baseline (i.e., attention UNet) is trained on 800 image samples while 5 shot (referring to a few-shot training using 5 samples) results for PMG Baseline is reported~\citep{xiao2021prior}. Similarly, for meta-learning approaches we provide results on 5, 10 and 20 shot episodic training. Test samples consist of only unseen ISIC data samples.}}
    \setlength{\tabcolsep}{6pt}
    \begin{tabular}{ l c c c c c } 
\toprule
        \textbf{Algorithm} & \textbf{K-shots} & \# \textbf{Tasks} & \textbf{Target Dataset} & \textbf{DSC}  \\ \midrule \midrule
  
    Naive Baseline & 800 & - & ISIC & 58.10  \\ \midrule

    \multirow{3}{*}{Semi-supervised. baseline~\citep{feyjie2020semi}} & 5 & - & ISIC & 61.38  \\
                                         & 10 & - & ISIC & 61.40  \\
                                         & 20 & - & ISIC & 60.79  \\ \midrule
                                         
    \multirow{1}{*}{PMG Baseline~\citep{xiao2021prior}} & 5 & - & ISIC & 67.00  \\ 
                                         \midrule                                      
    
    \multirow{3}{*}{Meta-learned (MAML)} & 5 & 50 & ISIC &  {75.62}\\
                                  & 10 & 50 & ISIC & {77.31} \\
                                  & 20 & 50 & ISIC & \textbf{79.60} \\ \bottomrule
                                  
     \multirow{3}{*}{Meta-learned (iMAML)} & 5 & 50 & ISIC &  \textbf{77.39}\\
                                  & 10 & 50 & ISIC & \textbf{79.17} \\
                                  & 20 & 50 & ISIC & \textbf{83.26} \\ \bottomrule                              
    \end{tabular}
    \label{tab2}

    \end{table*}

\subsection{Network Architecture}
Our proposed model architecture is shown in Fig.~\ref{fig:method}. Our figure is divided into stage 1 and stage 2.  In stage 1, meta-training is done on the support set, and the validation is done on the query set. Similarly, in stage 2, meta-testing is done on the test set. From the figure, we can observe that meta-training is performed as episodic tasks on two public datasets (PH\textsuperscript{2}~\citep{mendoncya2013dermoscopic}, and CVC-ClinicDB~\citep{{bernal2015wm}}). During the meta-testing stage (stage 2), an unseen task from the third dataset is provided with the optimized weights obtained from the first stage, \#1 and the gradient of the computed loss is used to readjust the final weights on only a few samples of this dataset (please refer to stage 2 part of Figure~\ref{fig:method}). The network consists of a sampler for creating support and query set for the few-shot setting of our experiment and for specific tasks. In all these settings, we use attention U-Net~\citep{oktay2018attention} as the meta learner to achieve segmentation maps. {The attention U-Net is used for each task's inner-level parameter optimization $\phi_i$.} We have a meta-gradient optimizer for computing the optimized weights fed to the attention U-Net {model}. Finally, the fine-tuned weight is used for the inference of the test samples, and the ground truth masks are predicted. 


\section{Experiments and Results} 
This section will describe the experimental setup, implementation details, and our results on each dataset. 

\subsection{Setup}{\label{sec:setup}}
\paragraph{\textbf{Experimental design.}}
All experiments in this work use few-shot supervised settings for which
$\textit{N}$-way, \textit{K}-shot tasks are randomly generated from two publicly available datasets. In this context, $\textit{N}$ refers to the number of classes and $\textit{K}$ refers to samples from each class. The number of classes $\textit{N}$ corresponds to the number of different data pools, making our experiments a 2-way $\textit{K}$-shot task. {Finally, the learned parameters were fine-tuned over an entirely new task drawn from the hold-out data pool for the meta-testing.} We present three sets of experiments: (i) tasks that comprised of samples exclusively from the \emph{Kvasir-SEG} (polyp) dataset or from the \emph{PH\textsuperscript{2}} (skin) dataset, (ii) tasks that are comprised of mixed samples, and (iii) tasks trained on the same class datasets and tested on {an entirely} different class, such as meta-training on skin datasets and meta-testing on polyp dataset.

\begin{figure}[t!]
    \centering
    \includegraphics[width=0.5\textwidth]{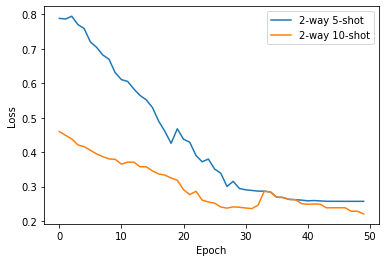}
    \caption{Comparison of learning curves between 2-way 5-shot  and 2-way 10-shot meta-learning {settings} that correspond to Table~\ref{tab2} for our proposed iMAML approach. Loss up to 50 epochs are provided to illustrate the convergence.}
    \label{fig:Loss}
\end{figure}

\paragraph{\textbf{Implementation details.}}
The meta-parameters were initialized with pre-trained weights from U-Net trained on brain \ac{MRI} scans~\citep{pedano2016radiology}. The meta-gradient is computed by applying conjugate gradient (CG), and the meta-parameters are updated using the Adam optimizer~\citep{Adam} with a learning rate of $10^{-5}$ and a weight decay of $0.0005$. Our convergence criteria is reached when the loss function does not change more than 0.001 over ten epochs. Figure~\ref{fig:Loss} shows the training convergence at {the} 50th epoch for {a} model trained in 2-way 5-shot and 2-way 10-shot settings. For the regularization of the computed learned weights, we fixed $\lambda=100$. The images and their corresponding ground truth were normalized in the range of [-1, 1] and resized to $256\times 256$. All implementations were done using the PyTorch framework, and experiments were conducted on NVIDIA Tesla V100-SXM3. 

\subsection{Results}
\begin{table*}[!t]
    \centering
        \caption{Episodic training on tasks comprised of both PH\textsuperscript{2} (skin) and Kvasir-SEG (polyp) instances. {This refers to our second experimental setup. Similar to Table~\ref{tab2}, here we present DSC metric scores for 5, 10, and 20 shots for meta-learning approaches again tested on unseen ISIC datasets.}}
    \setlength{\tabcolsep}{6pt}
    \begin{tabular}{ l c c c c c } 
    \toprule
        \textbf{Algorithm} & \textbf{K-shots} & \# \textbf{Tasks} & \textbf{Target Dataset} & \textbf{DSC}  \\ \midrule \midrule
  
    Naive Baseline & 800 & - & ISIC & 58.10  \\ \midrule
    
    \multirow{3}{*}{Semi-supv. Baseline~\citep{feyjie2020semi}} & 5 & - & ISIC & 61.38  \\ 
                                         & 10 & - & ISIC & 61.40  \\ 
                                         & 20 & - & ISIC & 60.79  \\ \midrule

      \multirow{1}{*}{PMG Baseline~\citep{xiao2021prior}} & 5 & - & ISIC & 67.00  \\ 
                                         \midrule      
                                         
       \multirow{3}{*}{Meta-learned (MAML)} & 5 & 50 & ISIC & {66.19}  \\ 
                                  & 10 & 50 & ISIC & {68.54} \\ 
                                  & 20 & 50 & ISIC & {70.61} \\ \bottomrule                                     
    \multirow{3}{*}{Meta-learned (iMAML)} & 5 & 50 & ISIC & \textbf{70.15}  \\ 
                                  & 10 & 50 & ISIC & \textbf{71.69} \\ 
                                  & 20 & 50 & ISIC & \textbf{72.48} \\ \bottomrule
    
    \end{tabular}
    \label{tab3}
\end{table*}

\begin{table*}[!t]
    \centering
        \caption{Episodic training on CVC-612 (polyp) and Kvasir-SEG (polyp) dataset. Here we provide quantitative results from our third experimental setup {(i.e., tasks comprising samples from two unique datasets of the same class)}. Similar to Table~\ref{tab2}, here we present DSC metric scores for 5, 10, and 20 shots for meta-learning approaches again tested on the unseen ISIC dataset.}
    \setlength{\tabcolsep}{6pt}
    \begin{tabular}{ l c c c c c } 
    \toprule
        \textbf{Algorithm} & \textbf{K-shots} & \# \textbf{Tasks} & \textbf{Target Dataset} & \textbf{DSC}  \\ \midrule \midrule
  
    Naive Baseline & 800 & - & ISIC & 58.10  \\ \midrule
    
    \multirow{3}{*}{Semi-supv. Baseline~\citep{feyjie2020semi}} & 5 & - & ISIC & 61.38  \\ 
                                         & 10 & - & ISIC & 61.40  \\ 
                                         & 20 & - & ISIC & 60.79  \\ \midrule
                                         
    \multirow{1}{*}{PMG Baseline~\citep{xiao2021prior}} & 5 & - & ISIC & 67.00  \\  \hline
    
    \multirow{3}{*}{Meta-learned (MAML)} & 5 & 50 & ISIC & {59.70}   \\ 
                                  & 10 & 50 & ISIC & {62.43}  \\ 
                                  & 20 & 50 & ISIC & {64.70} \\ \bottomrule

    \multirow{3}{*}{Meta-learned (iMAML)} & 5 & 50 & ISIC & \textbf{63.56}   \\ 
                                  & 10 & 50 & ISIC & \textbf{65.09}  \\ 
                                  & 20 & 50 & ISIC & \textbf{66.71} \\ \bottomrule
    
    \end{tabular}
    \label{tab4}
\end{table*}

\begin{table*}[t!h!]
    \centering
    \caption{Episodic meta-training on Kvasir-SEG (polyp) and PH\textsuperscript{2} (skin) dataset from the second experimental setup { (i.e., tasks comprising mixed samples of two unique datasets). Meta-testing is done on instances from unseen KvasirCapsule-SEG (wireless capsule endoscopy polyp) dataset. Here, naive baseline (attention U-Net) is trained on KvasirCapsule-SEG using 44 samples (80\%) and tested on remaining samples (20\%) as done for other meta-learning approaches.}}
    \setlength{\tabcolsep}{6pt}
    \begin{tabular}{ l c c c c c } 
    \toprule
        \textbf{Algorithm} & \textbf{K-shots} & \# \textbf{Tasks} & \textbf{Target Dataset} & \textbf{DSC}  \\ \midrule \midrule
  
    Naive Baseline & 44 & - & KvasirCapsule-SEG &  16.23 \\ \midrule
       \multirow{3}{*}{Meta-learned (MAML)} & 5 & 50 &  KvasirCapsule-SEG & {53.33}  \\ 
                                  & 10 & 50 & KvasirCapsule-SEG & {56.10}  \\ 
                                  & 20 & 50 & KvasirCapsule-SEG & {58.47}\\\midrule
     \multirow{3}{*}{Meta-learned (iMAML)} & 5 & 50 &  KvasirCapsule-SEG & \textbf{56.39}  \\ 
                                  & 10 & 50 & KvasirCapsule-SEG & \textbf{59.34}  \\ 
                                  & 20 & 50 & KvasirCapsule-SEG & \textbf{61.28}\\ \bottomrule
    \end{tabular}
    \label{tab5}
\end{table*}

\begin{table*}[t!h!]
    \centering
    \caption{Episodic meta-training on ISIC (skin) and PH\textsuperscript{2} (skin) {datasets} from the third experimental setup {(i.e., tasks comprising samples from two unique datasets of the same class). The meta-testing is done on instances from Kvasir-SEG and KvasirCapsule-SEG dataset. Here, both naive baseline attention-UNet, MAML and iMAML meta-learning approaches are compared.}}
    \setlength{\tabcolsep}{6pt}
    \begin{tabular}{ l c c c c c } 
    \toprule
        \textbf{Algorithm} & \textbf{K-shots} & \# \textbf{Tasks} & \textbf{Target Dataset} & \textbf{DSC}  \\ \midrule \midrule
  
    Naive Baseline & 800 & - & Kvasir-SEG &  60.53 \\ 
     & 44 & - & KvasirCapsule-SEG &  16.23 \\
        \midrule
    \multirow{3}{*}{Meta-learned (MAML)} & 5 & 50 & Kvasir-SEG & 59.30 \\ 
                                  & 10 & 50 & Kvasir-SEG & 61.72 \\ 
                                  & 20 & 50 & Kvasir-SEG & 64.09\\ \midrule
    \multirow{3}{*}{Meta-learned (iMAML)} & 5 & 50 & Kvasir-SEG & \textbf{62.00}  \\ 
                                  & 10 & 50 & Kvasir-SEG & \textbf{65.10}  \\ 
                                  & 20 & 50 & Kvasir-SEG & \textbf{66.58}\\ \midrule
\multirow{3}{*}{Meta-learned (MAML)} & 5 & 50 & KvasirCapsule-SEG& {52.26}  \\ 
                                  & 10 & 50 & KvasirCapsule-SEG & {54.09}  \\ 
                                  & 20 & 50 & KvasirCapsule-SEG & {57.47}\\ \midrule
    \multirow{3}{*}{Meta-learned (iMAML)} & 5 & 50 & KvasirCapsule-SEG& \textbf{53.80}  \\ 
                                  & 10 & 50 & KvasirCapsule-SEG & \textbf{55.35}  \\ 
                                  & 20 & 50 & KvasirCapsule-SEG & \textbf{58.19}\\ \bottomrule
    \end{tabular}
    \label{tab6}
\end{table*}

\begin{table*}[t!h!]
    \centering
    \caption{{Quantitative results on the study of the effect of Lovász  extension compared to the standard dice loss function in a meta-learning setting with 5 shot 2 way. The meta-training was done on two datasets (i.e., 2- way) namely CVC-612 and PH\textsuperscript{2}, and tested on unseen ISIC dataset.}}
    \setlength{\tabcolsep}{6pt}
    \begin{tabular}{ l c c c c c }
    \toprule
        \textbf{Algorithm} & \textbf{K-shots} & \# \textbf{Tasks} & \textbf{Target Dataset} & \textbf{DSC}  \\ \midrule \midrule
        Dice Loss  & 5 & 20 & ISIC & 73.90 \\
        Log(cosh(Dice Loss)) & 5 & 20 & ISIC & \textbf{76.85}   \\ \bottomrule
    \end{tabular}
    \label{tab7}
\end{table*}
We present results for three different experimental setups to illustrate the model efficacy compared to naive supervised attention U-Net and two recent SOTA few-shot methods used for medical image segmentation.

\begin{figure*} [!t]
    \centering
    \includegraphics[width=0.90\textwidth]{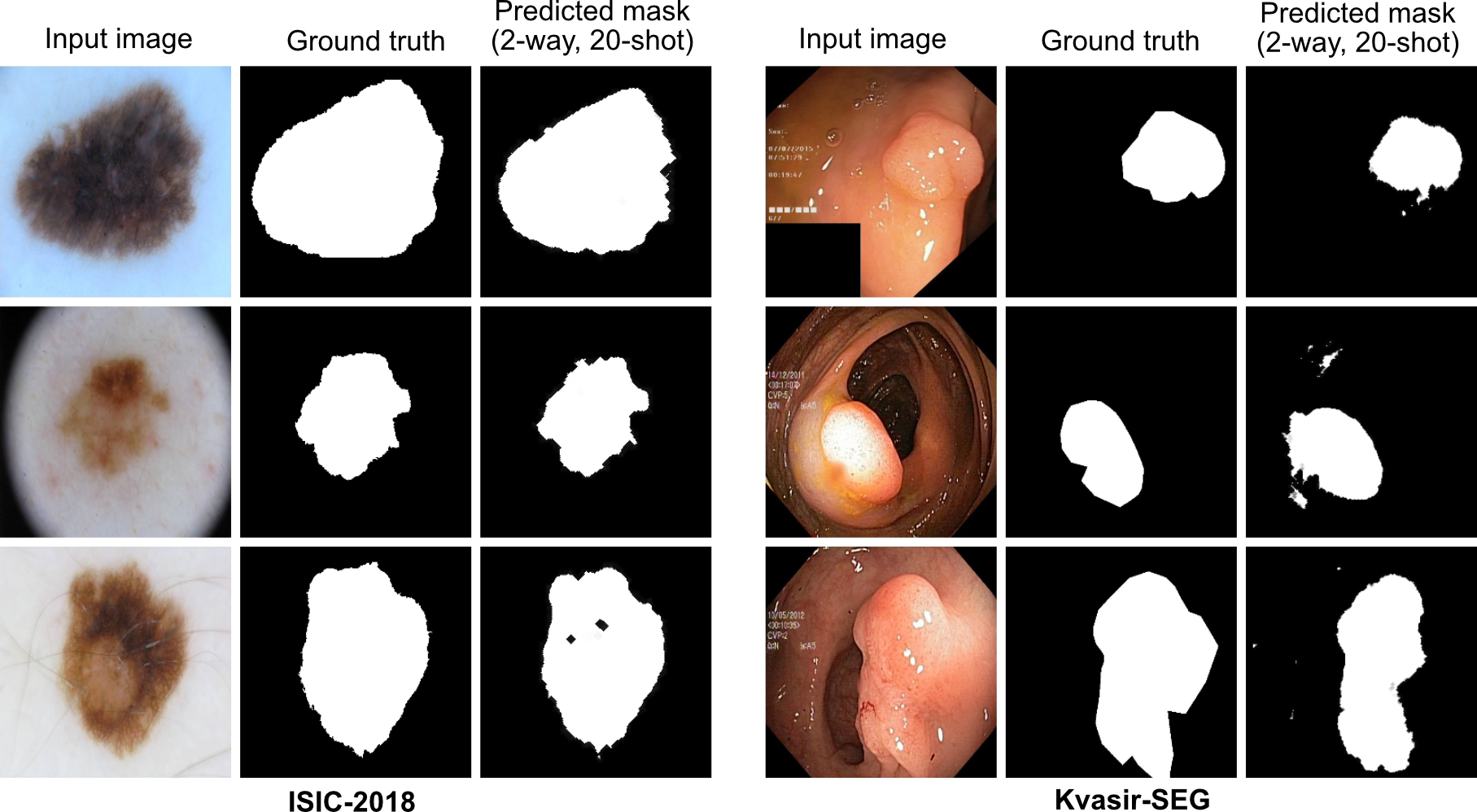}
     \caption{Qualitative results of the proposed method on ISIC-2018~\citep{codella2019skin,tschandl2018ham10000} {(left)} presented in Table~\ref{tab2} and {results on} Kvasir-SEG~\citep{jha2020kvasir} {(right) corresponding to Table~\ref{tab6}. (Left) represents our first experimental configuration, i.e., training with samples drawn uniquely from two datasets from two different class categories (in this case PH$^2$ and Kvasir-SEG)} while (right) corresponds to our third setup where samples comprise of two different datasets but of unique class (only skin datasets in this case).}
    \label{fig:qualitative}
\end{figure*}

\paragraph{\textbf{1. Meta-training with samples drawn exclusively from two unique datasets and unique categories:}}
Table~\ref{tab2} presents the episodic training of our meta-learning approach on the PH\textsuperscript{2} and Kvasir-SEG datasets consisting of skin and polyp categories, respectively. It can be observed that on the unseen ISIC dataset for test, our proposed iMAML-based segmentation outperformed the naive baseline U-Net by a very large margin of 25\% and by nearly 23\% and 16\% on the dice coefficient compared to the baseline semi-supervised method~\citep{feyjie2020semi} and the recent mask guided few-shot segmentation approach (PMG baseline)~\citep{xiao2021prior}, respectively. The qualitative results (Figure~\ref{fig:qualitative}, left) also provide insight that our method provided optimal segmentation masks for different skin lesion types. The proposed meta-learning-based segmentation obtained the highest dice coefficient of 77.39\%, 79.17\% and 83.26\%  for different $K$-shots, {i.e.,}  5, 10, and 20 shots, respectively.

\paragraph{\textbf{2. Tasks comprising mixed samples of two unique datasets:}}
Table~\ref{tab3} presents quantitative results for a different setting where the samples are mixed from two datasets (PH\textsuperscript{2} and Kvasir-SEG). Clearly, there is evidence of a performance drop in our meta-learning method. Nevertheless, the proposed algorithm consistently outperformed baseline methods. The best dice score of 72.48\% is obtained on the ISIC (skin) dataset under 2-way 20-shot setting, which is nearly 11.69\% and 5.48\% on the dice coefficient compared to the baseline semi-supervised method and PMG baseline model, respectively. Similarly, under this experimental setup, the segmentation results on the KvasirCapsule-SEG dataset using the iMAML and MAML algorithms are also captured in Table~\ref{tab5}. It can be observed that both the meta-learning algorithms \ac{iMAML} and \ac{MAML}, outperforms the baseline models by 45\% and 42\%, respectively, that was naively trained under classical supervised setting with limited 44 images that were available for {the} KvasirCapsule-SEG dataset.

%

\paragraph{\textbf{3. Tasks comprising samples from two unique datasets of the same class:}}
Table~\ref{tab4} and Table~\ref{tab6} represent meta-training on two unique datasets, but with the same categories and tested on a different class dataset. The categories here refers to a particular disease type (polyp or skin lesion). It can be observed that for episodic training conducted on the polyp datasets (CVC-ClinicDB and Kvasir-SEG) and tested on the skin dataset (see Table~\ref{tab4}), our method is still able to generalize better than the naive baseline approach trained on 800 samples and the recent semi-supervised approach. The best dice score of 66.71\% is obtained on {the} ISIC (skin) dataset under {a} 2-way 20-shot setting which is better by nearly 5.92\% compared to the baseline semi-supervised method and competitive to the PMG baseline. Similar observations can be found when the method is trained on the skin datasets such as  ISIC-2018 and PH\textsuperscript{2} datasets and tested on the Kvasir-SEG and KvasirCapsule-SEG polyp segmentation datasets (please refer to Figure~\ref{fig:qualitative} and Table~\ref{tab6}).

Additionally, we further investigated the effect of Lov{\'a}sz extension and standard dice loss function in a meta-learning setting using our first episodic training setup. Based on the experimental results  (see Table~\ref{tab7}), Lov\`{a}sz extension was chosen, which improved the segmentation result by nearly 3.00\%. 

\section{Discussion}
Owing to the challenges such as data scarcity and data mismatch in the medical field while applying deep learning techniques, the generalization capacity of the trained model is reduced during deployment. Furthermore, various biases are introduced during the data collection process that can induce data shift at test time and {derail the trained model's performance during a clinical deployment}. 

Acknowledging these challenges, the \ac{ML} community has carried out some studies, which includes {the work} by~\citet{feyjie2020semi} which is based on {a} semi-supervised {few-shot} learning method. Similarly, work by~\citet{dou2019domain} uses the meta-learning method MAML to tackle the challenge of data shifts due to various data sources. However, these previous works have not been tested for completely different anatomies and under {a} few shot settings. We propose a meta-learning method with {an} implicit gradient (iMAML) to overcome these challenges under {a} few shot settings.
The adopted meta-learning method is model agnostic and can take any other segmentation network as the meta-learner to learn the segmentation mask. For our experiment, we select the most popular segmentation network{, U-Net,} with an attention module as the meta-learner. The two SOTA methods used in comparison use few-shot learning approaches and hence can be directly compared. Adding any other supervised models would direct us to {similar} accuracy gains when used in a meta-learning framework. To test the efficacy of the iMAML algorithm, we arranged three different experiment setups (see Section~\ref{sec:setup}).
For carrying out the experiments, two datasets for skin, two datasets of normal colonoscopy and a dataset from video capsule endoscopy, which is a different {modality,} were used. The idea was to perform meta-training with tasks that {are} comprised of instances either from the same medical categories or different medical categories; to observe the generalization capacity of the algorithm. So, we picked two datasets that provided enough variability for episodic training. The results in Table~\ref{tab2} are from the first {experimental} setup where tasks are homogeneously comprised either only from the PH\textsuperscript{2} (skin) dataset or from the Kvasir-SEG (polyp) dataset and then tested on {the} ISIC dataset. The segmentation results from the \ac{iMAML} algorithm outperformed all the baseline models with the {largest} improvement of over 25\% compared to the naive baseline model. Furthermore, iMAML has an improvement of nearly 2\%-4\% over the standard MAML approach. 
Similarly, the results from the second experiment are tabulated in Table~\ref{tab3} where the meta-learning algorithm is trained on tasks comprised of both the PH\textsuperscript{2} (skin) and the Kvasir-SEG (polyp) datasets together. The segmentation performance on the test task from the ISIC (skin) dataset shows that the iMAML algorithm {outperforms} the naive baseline model by nearly 15\% and shows distinct performance gains over all other methods in our comparison. The overall degradation in performance of both meta-learning algorithms compared to the previous setup (see Table~\ref{tab2}) can be due to the increased variability in samples presented during the episodic meta-training that can make the network difficult to converge optimally to two different dataset attributes. 

The third experiment setup aims to test the generalization capacity of the meta-learning model on an entirely never seen task. The task is comprised of polyp datasets, {namely} from CVC-ClinicDB and Kvasir-SEG. Table~\ref{tab4} depicts the result of the third experiment setup where the performance of the meta-learning algorithm is further degraded. Again, this could be because of training on a completely different dataset acquired from a different device and a different class category. {Thus,} the proposed iMAML generalizes well and provides an improved result compared to naive baseline by 8.61\% and still eclipses the semi-supervised baseline method~\citep{feyjie2020semi} by 5.92\%. We further compared the test results between Kvasir-SEG and KvasirCapsule-Seg while training on tasks comprised of skin datasets only. The results captured in Table~\ref{tab5} demonstrate that even with datasets with few examples like KvasirCapsule-SEG, the meta-learning algorithms can perform better than the naive baseline models.

From the empirical observations, we can note that the DSC score is higher with ISIC as {the} target dataset in comparison with KvasirCapsule-SEG. This is {because} Kvasir-SEG and Kvasir Capsule datasets come from colorectal inspection (inside body), where the obtained images are often specular and have variable contrast based on their location. In contrast, {the} ISIC dataset is obtained from {dermatoscopy}, which is usually taken from the exposed skin {region} with polarised or non-polarised light sources and are concentrated closer to the area of interest and often have diffused reflection.

Empirically, we showed that the \ac{iMAML} algorithm {could} efficiently handle tasks with higher variations of instances during deployment.
The method is model agnostic and should be replicable with other imaging modalities. However, we have chosen skin and polyp datasets as the domain shift in these data are very observant due to 1) patient or population variability, 2) imaging type (e.g., colonoscopy vs capsule {endoscopy}) and 3) class and color variability in skin images (e.g., PH2 and ISIC).  It also illustrates that iMAML can be applied effectively in a {complex} problem like segmentation. During each of the experiment setups, the performance of the meta-learning algorithm is further improved by increasing the number of training tasks or the number of instances in each {task} which was a trade-off between training time and accuracy.  This provides a robust method to handle data scarcity {problems} while training a deep neural network.

The findings of the empirical studies suggest that {optimization-based} meta-learning can alleviate the problem of data generalization and data scarcity which is prominent in the medical domain. We showed that the idea of meta-learning is a plausible concept that can benefit medical image segmentation under few-shot settings. In the future, we want to investigate how prior information about feature embedding from each task could be used to reduce the training time. 

\section{Conclusion}
\label{sec:conclusion}
We proposed a novel model-agnostic meta-learning segmentation method in a few-shot setting that uses an implicit gradient-based optimization technique for improved model parameter estimation and generalization over unseen datasets with unique and seen categories. The proposed method improved performance and generalization capabilities compared to naive supervised techniques and the most recent few-shot segmentation approaches. We also demonstrated that the iMAML algorithm {performs} better {than a} popular meta-learning approach, MAML. Our {method} allowed the exploitation of available medical imaging datasets for training {such that the trained model can be applied on an} unseen dataset without requiring ample ground truth labels. Thus, the proposed method eliminates {the need for} abundant data for each specialized medical imaging category. However, the adopted meta-learning algorithm (iMAML) showed only marginal performance gain when trained with tasks comprised of instances from various medical categories. The generalization capacity of the \ac{iMAML} algorithm is reduced when trained with skewed tasks, for example, tasks comprising of instances from skin and polyp datasets. {To address such an issue, we will aim to shuffle channels or mix embedded features between instances of datasets while performing meta-training in future work.} Nevertheless, this meta-learning approach could potentially contribute to developing clinically {deployable} systems for real-world application in the future.


%
\section*{Acknowledgment}
D. Jha is funded by the PRIVATON project (\#263248) which is funded by Research Council of Norway (RCN).  S. Ali is supported by the National Institute for Health Research (NIHR) Oxford Biomedical Research Centre (BRC). Our experiments were performed on the Experimental Infrastructure for Exploration of Exascale Computing (eX3) system, which is financially supported by RCN under contract 270053. The views expressed are those of the author(s) and not necessarily those of the NHS, the NIHR or the Department of Health. 

\section*{Author contribution}
RK, DJ and SA wrote most of the manuscript with input from all the authors. RK conducted all the experiments reported in the paper. RK and SA revised the manuscript with valuable feedback for corrections from DJ, SH, VT, MAR, and PH. All authors agreed for submission.

\bibliographystyle{model2-names.bst}\biboptions{authoryear}
\bibliography{refs}
\end{document}